\documentclass[10pt,twocolumn,letterpaper]{article}

\usepackage[margin=1in]{geometry}
\usepackage[T1]{fontenc}
\usepackage[utf8]{inputenc}

\usepackage{mathpazo}

\usepackage[utf8]{inputenc} 
\usepackage[T1]{fontenc}    
\usepackage{url}            
\usepackage{booktabs}       
\usepackage{amsfonts}       
\usepackage{nicefrac}       
\usepackage{microtype}      
\usepackage{cvpr}              
\usepackage{mathtools,bbm}
\usepackage{color}
\usepackage{multirow}
\usepackage{caption}
\usepackage{subcaption}
\usepackage{makecell}
\usepackage{wrapfig}
\usepackage{textcomp}
\usepackage{gensymb}
\usepackage{marginnote}
\usepackage{algorithm}
\usepackage{algorithmic}
\usepackage{array}

\definecolor{cvprblue}{rgb}{0.21,0.49,0.74}
\usepackage[pagebackref,breaklinks,colorlinks,allcolors=cvprblue]{hyperref}

\def\paperID{9} 
\def\confName{CVPR}
\def\confYear{2025}

\title{Efficient LLaMA-3.2-Vision \\ 
by Trimming Cross-attended Visual Features}

\author{
\begin{tabular}{@{}c c}
Jewon Lee$^{1}$\quad
Ki-Ung Song$^{1}$\quad
Seungmin Yang$^{1}$\quad
Donguk Lim$^{1}$\quad\\
Jaeyeon Kim$^{1}$\quad
Wooksu Shin$^{1}$\quad
Bo-Kyeong Kim$^{1}$\quad
Yong Jae Lee$^{2}$\quad
Tae-Ho Kim$^{1}$\thanks{Corresponding author.}\vspace{6pt} 
\end{tabular}\\
$^{1}$Nota Inc.\quad$^{2}$University of Wisconsin-Madison\\
\texttt{\scalebox{0.765}{ \{jewon.lee, thkim\}@nota.ai}}
}

\begin{document}
\maketitle
\begin{abstract}
 Visual token reduction lowers inference costs caused by extensive image features in large vision-language models (LVLMs). Unlike relevant studies that prune tokens in self-attention-only LVLMs, our work uniquely addresses cross-attention-based models, which achieve superior performance. We identify that the key-value (KV) cache size for image tokens in cross-attention layers significantly exceeds that of text tokens in self-attention layers, posing a major compute bottleneck. To mitigate this issue, we exploit the sparse nature in cross-attention maps to selectively prune redundant visual features. Our \textbf{Trimmed Llama} effectively reduces KV cache demands {without requiring additional training}. By benefiting from 50\%-reduced visual features, our model can reduce inference latency and memory usage while achieving benchmark parity.
\end{abstract}

\section{Introduction}
\begin{figure}[h]
    \centering
    \includegraphics[width=0.43\textwidth]{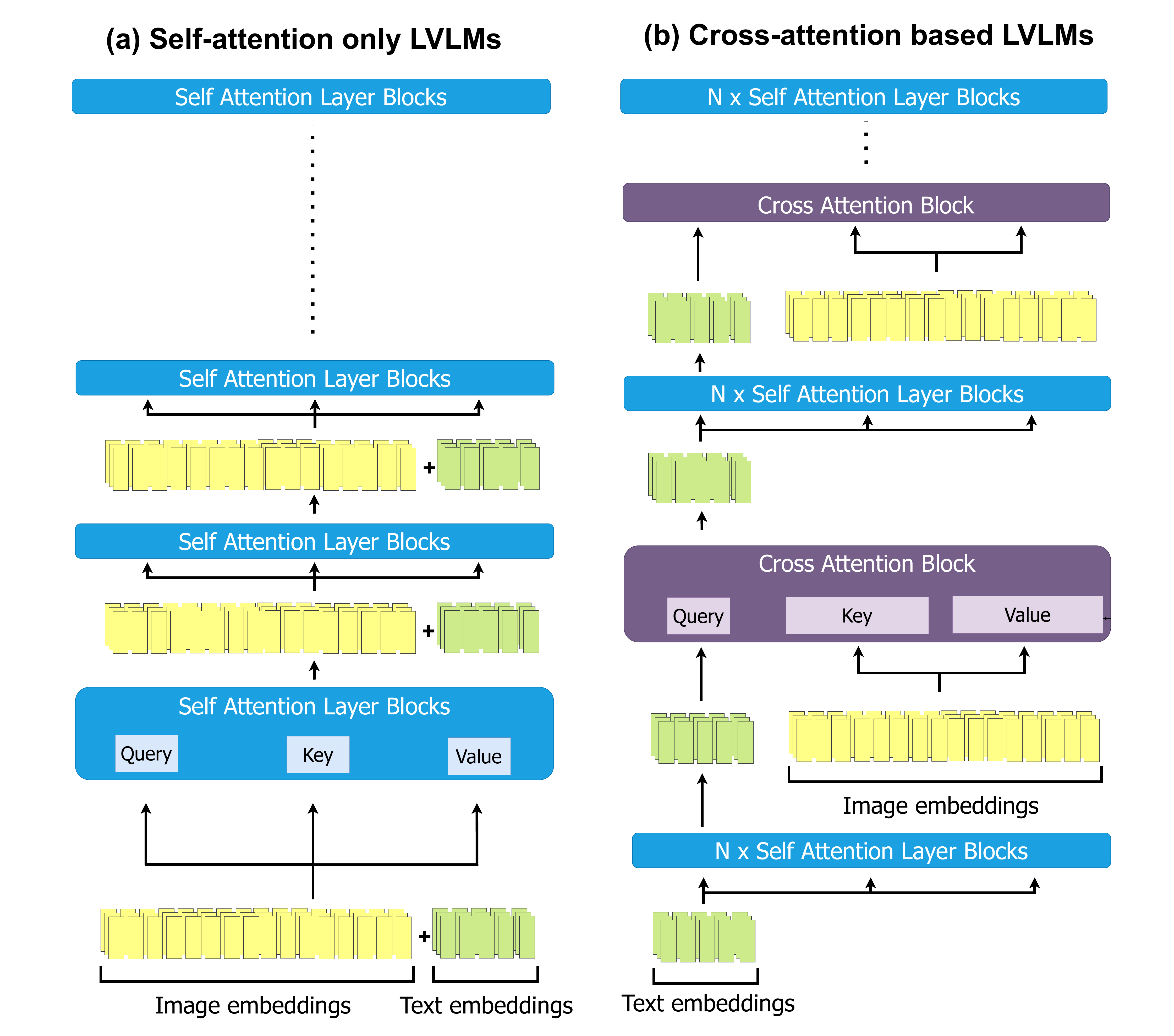}
    \caption{{\textbf{Comparison of LVLM architectures.} (a) Self-attention-only models process both image and text embeddings in all attention layers. (b) Cross-attention-based models use image features exclusively for KV operations in cross-attention layers, enabling efficient multimodal integration.}}
    \vspace{-1.3em}
    \label{fig_comp_sa_ca}
\end{figure}

\vspace{-0.15em}
\label{sec:intro}
Large Vision Language Models (LVLMs), such as LLaVA \citep{liu2024visual}, commonly utilize self-attention-only architectures in their large language models (LLMs). These models process visual inputs as sequences of hundreds or thousands of tokens \citep{li2024llava, chen2024expanding} alongside textual prompts (see Figure \ref{fig_comp_sa_ca}(a)). However, their computational complexity grows quadratically with input length, limiting deployment in high-resolution or feature-rich environments.

In contrast, cross-attention-based architectures, exemplified by Flamingo \citep{alayrac2022flamingo}, integrate visual features into LLMs as key-value (KV) pairs in text-visual attention computations (see Figure \ref{fig_comp_sa_ca}(b)). This design achieves linear computational scaling for image processing, enabling efficient processing of visual inputs. Recent advancements, such as Llama-3.2-Vision \citep{dubey2024llama, llama3.2-11b-vision-instruct}, demonstrate their capability, positioning them as robust alternatives to self-attention-only models.

Enhancing LVLM efficiency involves optimizing vision token computations by exploiting the characteristics of causal self-attention \citep{xiao2023efficient, ge2023model, chen2025image, liu2024efficient, he2024zipvl}. However, the study of efficient text-visual cross-attention mechanisms has not been thoroughly investigated.

\begin{figure}[t]
\centering
\includegraphics[width=1.0\linewidth]{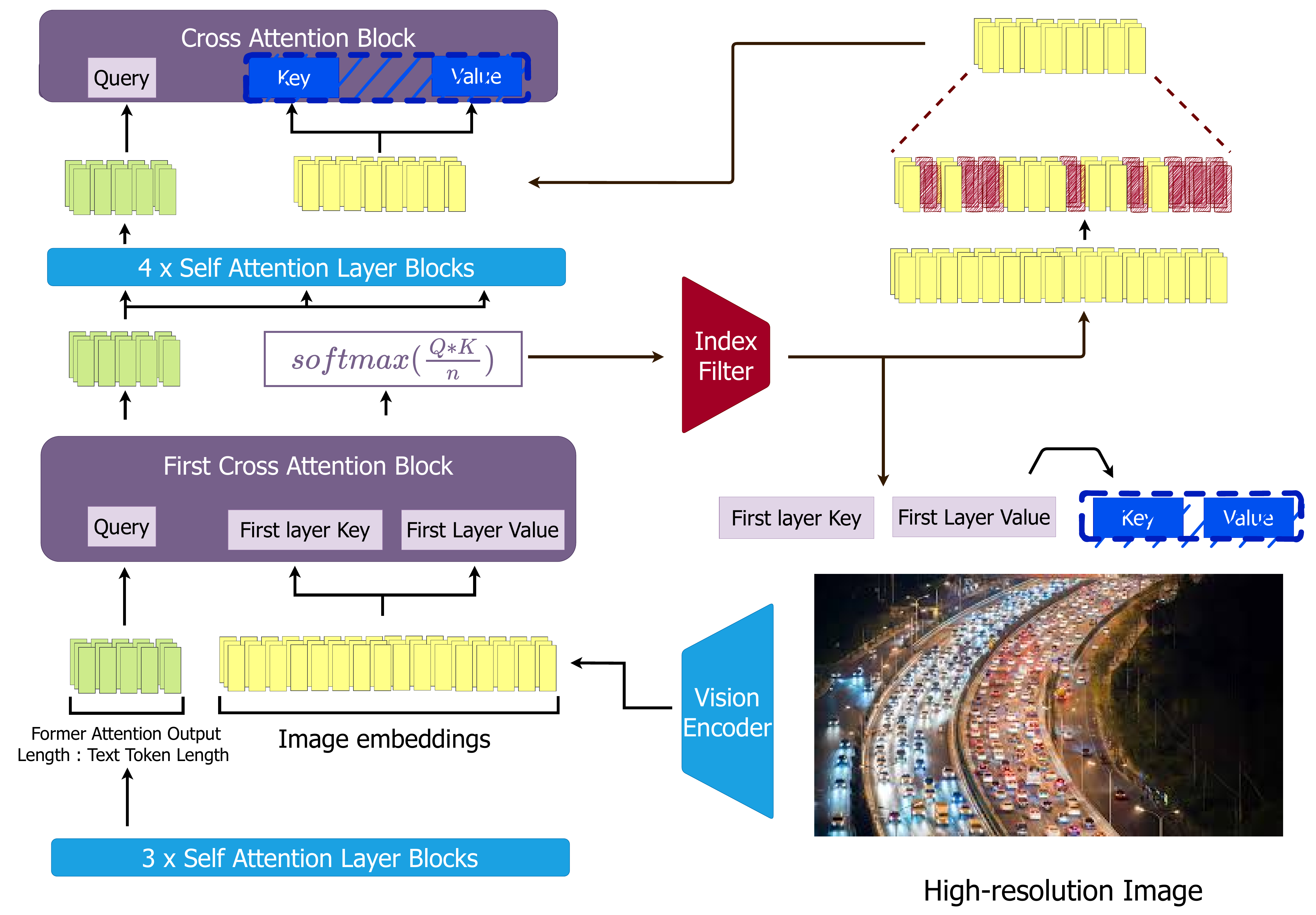}

\vspace*{-0.07in}
\caption{\textbf{Proposed method.} Image features are pruned in the first cross-attention block using a criterion derived from attention weights. The features serve as inputs for the keys and values in subsequent cross-attention layers, with the compressed keys and values stored in the KV cache (blue-shaded area).} 
\label{fig:compression-overview}
\vspace{-0.4em}
\end{figure}

In this work, we uncover the sparsity in cross-attention {maps} of LVLMs, revealing a consistent layer-wise pattern where the majority of visual features are selected in earlier layers, with minimal variation in subsequent layers. With these insights, we propose a novel method named \textbf{Trimmed Llama} that leverages the \textit{sparsity} and \textit{inter-layer resemblance of cross-attention patterns} to trim out redundant image features during inference (see Figure \ref{fig:compression-overview}). Our approach is training-free and can achieve a minimal performance trade-off while reducing KV cache budget and computation cost, resulting in efficient inference. Our contributions are summarized as follows.

\begin{enumerate}[itemsep=0em]

\item[$\circ$] Discovery of sparsity in cross-attention: Unlike prior work solely focusing on self-attention-only LVLMs, we target recent cross-attention-based models. We identify that visual attention mechanisms in different cross-attention layers exhibit a shared sparse pattern.

\item[$\circ$] Novel visual token pruning method: Based on the observed sparsity, we leverage head-wise attention scores to filter out unimportant visual features, thereby reducing KV cache overhead.

\item[$\circ$] {Solid} empirical validation: 
We test our approach across diverse benchmarks, ranging from vision-based multiple-choice questions to image grounded open-ended generation task, achieving performance on par with the original model while utilizing only \textbf{50\%} of the image features.

\setlength{\leftskip}{0pt}
\end{enumerate}

\begin{figure}[t]
\centering
\includegraphics[width=1.0\linewidth]{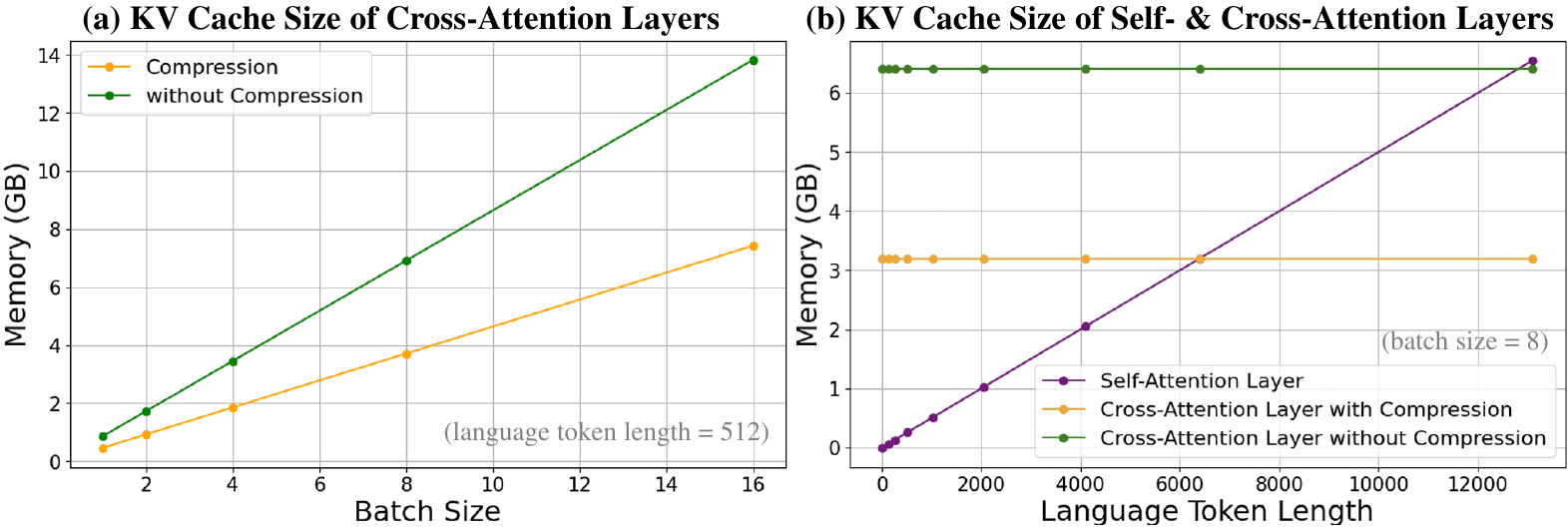}

\vspace*{-0.07in}
\caption{\textbf{KV cache memory.} (a) As batch size increases, the KV cache volume from image features grows. (b) As the language token count grows, the KV cache size in cross-attention still dominates that of self-attention, up to a certain number of tokens.} 
\label{fig:kv-memory}
\vspace{-0.25em}
\end{figure}

\section{Cross-attention Redundancy} \label{sec:cross-attention-analysis}
\vspace{-0.15em}

\subsection{{Motivation: Heavy Computation from Cross-Attention KV Cache}}

\noindent{\textbf{Benefit of Cross-Attention Layers in LVLMs.}} Higher image resolutions generally improve model {performance} \citep{lin2024vila, dai2024nvlm, xu2024llava} but result in more image tokens. This poses a computational challenge for self-attention-only {models \citep{liu2024visual, deitke2024molmo, internvl2-8b}}, whose complexity grows quadratically with token count. Cross-attention {architectures \citep{llama3.2-11b-vision-instruct, alayrac2022flamingo}}, by contrast, mitigate these issues by limiting the handling of image tokens to specific layers, avoiding quadratic scaling.

\noindent{\textbf{KV Cache of Cross-Attention Layers}. Though cross-attention layers in LVLMs improve efficiency, their KV caches are still heavy.} For Llama-3.2-11B-Vision-Instruct \citep{llama3.2-11b-vision-instruct} as our baseline, {visual} token length ranges from 1,601 tokens (e.g., 384×384 resolution) to 6,404 tokens (e.g., 720p, 1080p). Figure \ref{fig:kv-memory}(a) shows that the KV cache memory in cross-attention layers grows significantly with batch size. Figure \ref{fig:kv-memory}(b) {shows} {that} {the KV cache size from image features in cross-attention layers surpasses that from text features in self-attention layers, up to a certain number of language tokens. Moreover,} the cross-attention KV cache remains constant regardless of generation steps. This analysis emphasizes the cross-attention KV cache as a key bottleneck in model inference.

\subsection{{Insights from Structured and Sparse Cross-Attention Patterns}}

\label{cross-attention-experiment}
\begin{figure}[t]
    \centering
    \begin{subfigure}{0.95\columnwidth}
        \centering
        \includegraphics[width=\textwidth]{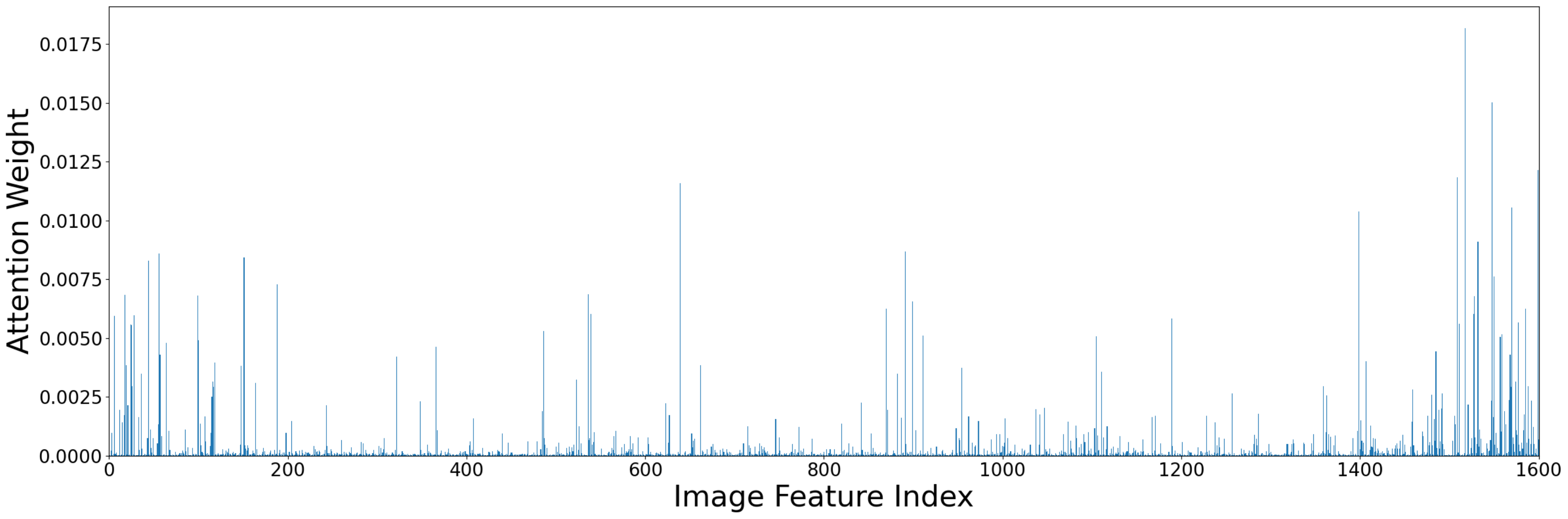}
        \vspace{-1.5em}
        \caption{Attention weights at the first cross-attention layer (x-axis: index of image features).} 
        \label{fig:img-level}
    \end{subfigure}    
    \vspace{-0.5em}
    \par\medskip
    \begin{subfigure}{0.95\columnwidth}
        \centering
        \includegraphics[width=\textwidth]{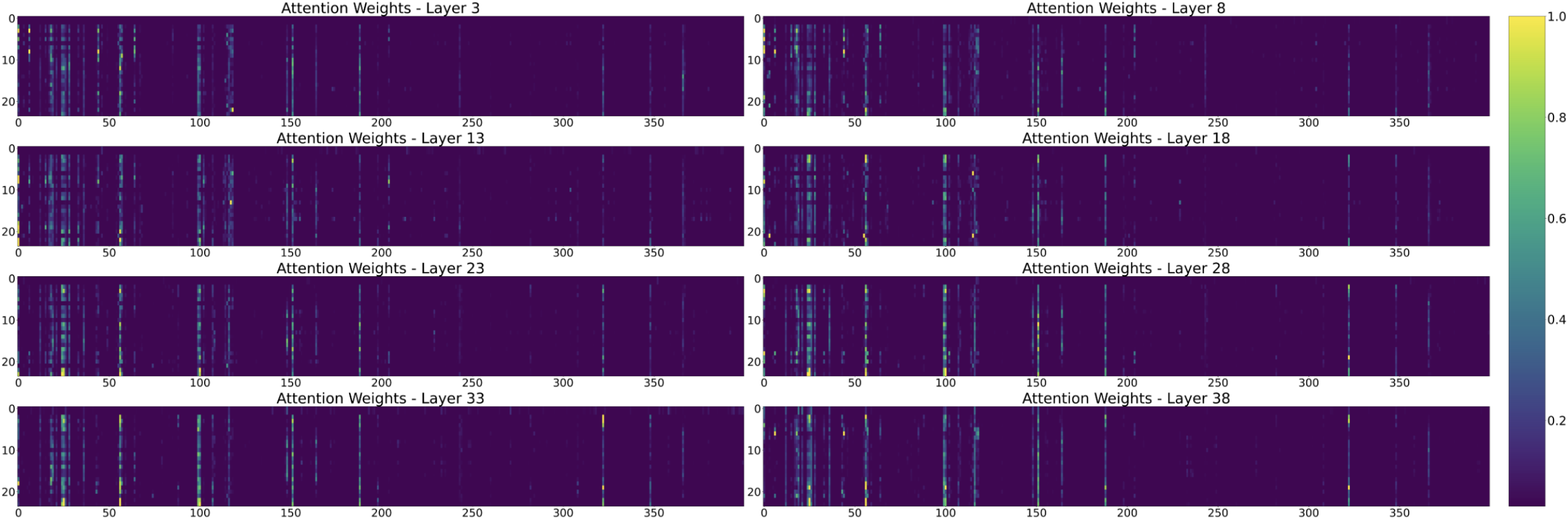}
        \vspace{-0.5em}
        \caption{Cross-attention weight patterns across different layers (x-axis: index of image features; y-axis: index of text query features).}
        \label{fig:cross-attn-pattern}        
    \end{subfigure}
    \vspace{-0.2em}
    \caption{\textbf{Aggregated cross-attention weights.} (a) The attention weights at the first cross-attention layer are summed over attention heads and text queries. (b) The attention weights for each cross-attention layer are summed over heads and visualized with the sequence length clipped to 400 for better visibility. Over different layers, specific image tokens consistently attract more attention from query tokens, indicating a structured sparse pattern.}
    \label{fig:cross-attention-visualization}
    \vspace{-1em}
\end{figure}

To investigate the processing of image tokens within the model, we analyze the attention patterns across multiple layers of the cross-attention mechanism. For a 384×384 image as the input of Llama-3.2-11B-Vision-Instruct, we aggregate the attention weights by summing them along two key dimensions, head-wise and query-wise. In Figure \ref{fig:img-level}, we observe that certain image tokens consistently attract attention from query tokens. This suggests that the model is selectively focusing on a relatively small subset of image features while ignoring others, indicating a potential mechanism for feature selection during cross-modal processing.

Figure \ref{fig:cross-attn-pattern} depicts the attention patterns across different cross-attention layers{, which reveal} two distinct phenomena. First, a vertically structured attention pattern is evident within each layer, signifying that attention weights are consistently allocated to the same indices, regardless of the specific language query. This behavior suggests the potential to identify globally salient indices. Second, there is an inter-layer consistency in attention patterns; the distribution of attention remains remarkably stable across successive layers. This lack of substantial variation implies that the model’s cross-attention mechanism might converge to a fixed pattern early, with minimal variation afterward. These findings contribute to our understanding of how cross-modal attention mechanisms operate, particularly in visual-linguistic tasks, and suggest avenues to optimize inference efficiency.
\section{Trimming Visual Features in Cross-Attention-Based LVLMs}
\begingroup
\renewcommand{\arraystretch}{1.10} 

\begin{table*}
    \centering  
    \resizebox{0.87\textwidth}{!}{  
        \begin{tabular}{>{\raggedright\arraybackslash}m{3.2cm}|
                        >{\centering\arraybackslash}m{2cm}|
                        >{\centering\arraybackslash}m{2cm}|
                        >{\centering\arraybackslash}m{2cm}|
                        >{\centering\arraybackslash}m{2cm}|
                        >{\centering\arraybackslash}m{2cm}|
                        >{\centering\arraybackslash}m{2cm}|
                        >{\centering\arraybackslash}m{2cm}}
            \specialrule{.2em}{.1em}{.1em} 
            {Model} & {Method} & SEED-Bench Image & MME & MME-cog. & MME-per. & MMVP & LLaVA-Bench \\ \hline
            \multirow{6}{*}{Llama-3.2-V Inst. 11B} & Original & 72.6 & 1685.9 & 307.1 & 1378.7 & 46.7 & 88.3 \\ \cline{2-8}
            & \multirow{2}{*}{$K_{\text{ratio}}=0.25$} & 72.3 & \textbf{1687.3} & \textbf{312.5} & 1374.8 & \textbf{47.3} & 88.1 \\ 
            & & \cellcolor[HTML]{C0C0C0} 61.5\% & \cellcolor[HTML]{C0C0C0} 65.1\% & \cellcolor[HTML]{C0C0C0} 65.1\% & \cellcolor[HTML]{C0C0C0} 65.1\%  & \cellcolor[HTML]{C0C0C0} 61.9\% & \cellcolor[HTML]{C0C0C0} 72.7\% \\ \cline{2-8}
            & \multirow{2}{*}{$K_{\text{ratio}}=0.20$} & 72.1 & 1682.8 & 307.9 & 1374.9 & \textbf{47.3} & 87.3 \\ 
            & &\cellcolor[HTML]{C0C0C0} 51.7\% &\cellcolor[HTML]{C0C0C0} 53.7\% &\cellcolor[HTML]{C0C0C0} 53.7\% &\cellcolor[HTML]{C0C0C0} 53.7\% &\cellcolor[HTML]{C0C0C0} 46.7\% &\cellcolor[HTML]{C0C0C0} 61.6\% \\ \cline{2-8}
            & \multirow{2}{*}{$K_{\text{ratio}}=0.15$} & 71.4 & 1669.1 & 305.7 & 1363.4 & 45.3 & \textbf{88.3} \\ 
            & &\cellcolor[HTML]{C0C0C0} 40.7\% &\cellcolor[HTML]{C0C0C0} 41.6\% &\cellcolor[HTML]{C0C0C0} 41.6\% &\cellcolor[HTML]{C0C0C0} 41.6\% &\cellcolor[HTML]{C0C0C0} 37.2\% &\cellcolor[HTML]{C0C0C0} 49.1\% \\ \cline{2-8}
            & \multirow{2}{*}{$K_{\text{ratio}}=0.10$} & 69.8 & 1675.8 & 297.86 & 1378.0 & 39.3 & 84.9 \\ 
            & &\cellcolor[HTML]{C0C0C0} 28.2\% &\cellcolor[HTML]{C0C0C0} 28.9\% &\cellcolor[HTML]{C0C0C0} 28.9\% &\cellcolor[HTML]{C0C0C0} 28.9\% &\cellcolor[HTML]{C0C0C0} 26.7\% &\cellcolor[HTML]{C0C0C0} 35.3\% \\ \cline{2-8}
            & \multirow{2}{*}{$K_{\text{ratio}}=0.05$} & 62.3 & 1586.1 & 297.5 & 1288.6 & 33.0 & 83.5 \\ 
            & &\cellcolor[HTML]{C0C0C0} 14.2\% &\cellcolor[HTML]{C0C0C0} 15.5\% &\cellcolor[HTML]{C0C0C0} 15.5\% &\cellcolor[HTML]{C0C0C0} 15.5\% &\cellcolor[HTML]{C0C0C0} 14.2\% &\cellcolor[HTML]{C0C0C0} 20.5\% \\ 
            \hline \hline
            \multirow{3}{*}{Llama-3.2-V Inst. 90B} & Original & 76.3 & 2029.2 & 423.9 & 1605.3 & 56.7 & 92.0 \\ \cline{2-8}
            & \multirow{2}{*}{$K_{\text{ratio}}=0.25$} & 75.9 & 2034.2 & \textbf{444.6} & 1589.6 & 54.7 & \textbf{93.9} \\ 
            & &\cellcolor[HTML]{C0C0C0} 74.2\% &\cellcolor[HTML]{C0C0C0} 71.6\% &\cellcolor[HTML]{C0C0C0} 71.6\% &\cellcolor[HTML]{C0C0C0} 71.6\% &\cellcolor[HTML]{C0C0C0} 71.3\% &\cellcolor[HTML]{C0C0C0} 75.2\% \\ \cline{2-8}
            & \multirow{2}{*}{$K_{\text{ratio}}=0.15$} & 75.4 & \textbf{2065.2} & 423.9 & \textbf{1641.3} & 56.6 & 92.2 \\ 
            & &\cellcolor[HTML]{C0C0C0} 51.0\% &\cellcolor[HTML]{C0C0C0} 48.4\% &\cellcolor[HTML]{C0C0C0} 48.4\% &\cellcolor[HTML]{C0C0C0} 48.4\% &\cellcolor[HTML]{C0C0C0} 45.7\% &\cellcolor[HTML]{C0C0C0} 52.4\% \\ 
            \specialrule{.2em}{.1em}{.1em} 
        \end{tabular}
    }
    \vspace{-0.2em}
    \caption{{\textbf{Performance of Llama-3.2-Vision-Instruct on various benchmarks.}} The value in grey {denotes} the mean percentage of {remaining} image features for each \(K_{ratio}\). Bold values {denote} performance comparable to or better than the full-cache baseline.}  \label{tab:llama3.2-11B-table}
    \vspace{-0.5em}
\end{table*}
\endgroup

With insights from Section \ref{cross-attention-experiment}, our {method} leverages \textit{head-wise} attention scores accumulated across language sequences to remove unimportant image features. For each attention head of the first cross-attention layer, the top-k most salient image features are identified based on their attention scores. Then, the union of these top-k sets across all heads is merged to determine the final selection of important features, ensuring a focused representation of the image.

Precisely, the sum of query-wise attention weights is computed for the cumulative importance score \(p_i^h = \sum_{j=0}^{m-1} \alpha_{i}^{j,h}\) (\(m\): input query tokens, \(i\): image feature index, \(j\): query token index, \(h\): head index, \(L\): set of image feature indices, \(H\): set of head indices, and \(\alpha_{i}^{j,h}\): attention score of image feature). The importance scores \(p_i^h\) are aggregated into \(P_h\), where \(p_i^h \in P_h\), \(\forall i \in L, \forall h \in H\).

Each image feature \(f_i^h\) at head \(h\) belongs to \(\mathbb{F}_h = \{f_i^h \mid i = 0, 1, 2, \dots, |L|-1\}\). Critical tokens are selected by evaluating the importance of image tokens uniquely per head, such that \(\mathbb{T}_h = \{i \mid p_i^h \in \text{top-k}(P_h, \text{top-k} = K_{ratio} \cdot |L|)\}\). The selected features for each head are \(\mathbb{F}_{select}^h = \{f_i^h \mid \mathbbm{1}_{\mathbb{T}_h} = 1\}\), and the total set of selected feature is \(\mathbb{F}_{select} = \bigcup_{h \in H} \mathbb{F}_{select}^h\). Here, \(K_{ratio}\) is an input parameter that determines the fraction of the feature space selected based on top-k criteria for each attention head.
\section{Experimental Setup}
\vspace{-0.15em}

\noindent \textbf{Models.}  We used the Llama-3.2-Vision-\{11B, 90B\}-Instruct models \citep{dubey2024llama} in our experiments. Compared to other cross-attention-based LVLMs like Open-Flamingo \citep{awadalla2023openflamingo} and Otter \citep{li2023mimic}, the Llama-3.2-Vision family exhibits superior capabilities by leveraging a significantly larger amount of visual tokens. 
Due to the limited visual tokens and lower performance of earlier models, we focused on the recent Llama-3.2-Vision for better compression results.

\noindent \textbf{Benchmark Datasets.} We used various benchmark datasets \citep{zhang2024lmmsevalrealitycheckevaluation} to assess its performance in vision-language tasks. Specifically, we {conducted} experiments on MMVP \citep{tong2024eyes} for binary classification question answering focusing on CLIP \citep{radford2021learning} blind pairs, MME \citep{fu2024mmecomprehensiveevaluationbenchmark} for fine-grained task-driven benchmark, SEED-Bench \citep{li2023seed} as vision-grounded multiple choice question answering benchmark and LLaVA-Bench \citep{liu2024visual} for open-ended vision-grounded generation.

\vspace{-0.1em}
\section{Results}
\vspace{-0.2em}

\begin{table}[]
    \centering
    \renewcommand{\arraystretch}{1.5} 
    \resizebox{1.0\linewidth}{!}{  
        {\Huge 
        \begin{tabular}{c|c|c|c|c|c}
            \specialrule{.2em}{.1em}{.1em}
            Feature Util. & Batch 1 & Batch 4 & Batch 8 & Batch 16 & Batch 32 \\ 
            \hline
            100\% (Orig.)         & 95.1ms    & 358.4ms  & 751.7ms  & 1648.6ms  & 3940.0ms  \\ \hline \rowcolor[HTML]{ECF4FF} 
            50.9\%                  & 91.2ms (4.1\%) & 332.9ms (7.1\%)  & 660.8ms (12.1\%) & 1414.7ms (14.2\%) & 3165.5ms (19.7\%) 
            
            \\ \rowcolor[HTML]{ECF4FF}  
            \hline
            39.6\%                  & 91.0ms (4.3\%) & 317.5ms (11.4\%) & 646.3ms (14.0\%) & 1347.7ms (18.3\%) & 2916.7ms (26.0\%) \\ 
            
            \specialrule{.2em}{.1em}{.1em}
        \end{tabular}
        }
    }
    \vspace{-0.2em}
    \caption{\textbf{Inference latency of the backbone LLM evaluated across different feature utilization ratios.} Parenthetical values indicate the relative latency reduction compared to the baseline model. The experiment was conducted using  Llama-3.2-11B-Vision-Instruct on an A100 80GB GPU.}
    \vspace{-0.2em}
    \label{tab:time_table}
\end{table}

\begin{figure}[t]
  \centering
    \includegraphics[width=\linewidth]{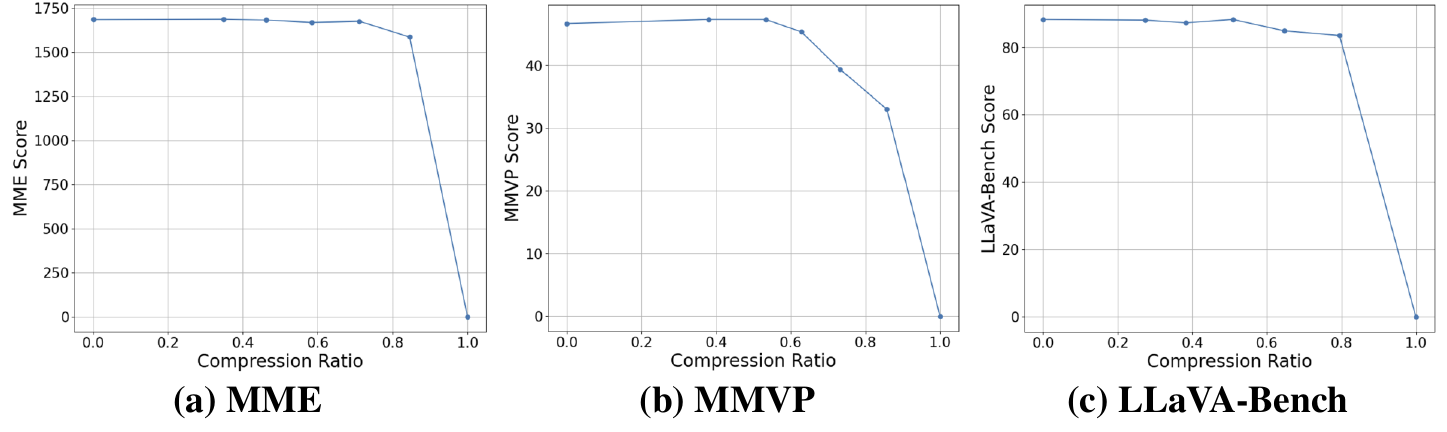}    
\vspace{-1em}
  \caption{\textbf{Results under different compression ratios.} Even with up to 50\% reduction of visual features, our method retains the performance of the original model.}
    \label{fig:score_by_ratio}
\end{figure}

\begin{figure}[t]
  \centering
    \includegraphics[width=0.95\linewidth]{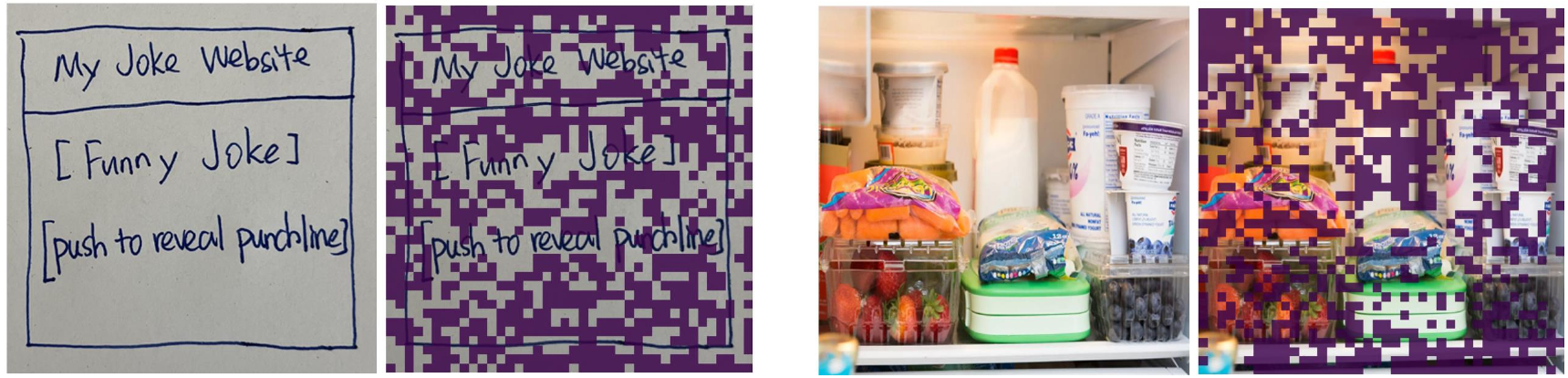}    
\vspace*{-0.05in}
  \caption{\textbf{Visualization of compression.} Purple patches indicate features trimmed by our method.}
  
  \label{fig:visualization}
\end{figure}

\noindent {\textbf{Main Results.} Table \ref{tab:llama3.2-11B-table} demonstrates} that our method consistently outperforms or achieves comparable performance {while} leveraging 40\(\sim\)50\% of the image features. Notably, the pruning ratios are \textit{adaptively allocated} for each task, as evidenced by LLaVA-Bench, an open-ended generation task {utilizing} more image features compared to other benchmarks. Figure \ref{fig:score_by_ratio} {shows} that {our approach} effectively maintains performance across benchmarks, even as the compression ratio increases. {Figure \ref{fig:visualization} shows that our method effectively preserves salient visual information (e.g., text cues or everyday objects) while pruning unimportant features.}

\noindent \textbf{Latency Reduction.}
Table \ref{tab:time_table} shows the inference speedup for the first token when utilizing 40\(\sim\)50\% of the image features. Our method reduces latency by pruning key and value inputs in the cross-attention layers. Since image features are pruned after the first cross-attention layer, both the key-value projections and the attention operations are consequently reduced. Furthermore, the impact of the reduction grows more significant with larger batch sizes.

\noindent \textbf{KV Cache Memory Reduction.} By removing image features after the initial cross-attention layer, we achieve optimal computational efficiency in reducing FLOPs. Figure \ref{fig:kv-memory} shows the impact of our approach on KV cache memory {(indicated with `Compression'). The amount of reduced cache size is amplified with larger batch sizes, highlighting the efficiency of our method under high-throughput conditions.}

\begin{table}[]
    \centering
    \resizebox{1.0\linewidth}{!}{ 
        \begin{tabular}{c|c|c|c|c}
            \specialrule{.2em}{.1em}{.1em}
            \rowcolor[HTML]{FFFFFF} 
            Method & SEED-Bench Image & MME & MMVP & LLaVA-Bench \\ \hline
            \multirow{2}{*}{Ours (budget < 0.50)} & 71.4 & 1669.1 & 47.3 & 88.3 \\ \cline{2-5} 
            
            & \cellcolor[HTML]{C0C0C0}40.7\% & \cellcolor[HTML]{C0C0C0}41.6\% & \cellcolor[HTML]{C0C0C0}46.7\% & \cellcolor[HTML]{C0C0C0}49.1\% \\ \hline
            Random (0.50) & 67.00 & 1537.1 & 44.7 & 83.2 \\ \hline
            Spatial (0.50) & 71.8 & 1627.7 & 46.0 & 85.9 \\ 
            \specialrule{.2em}{.1em}{.1em}
        \end{tabular}
    }
        \vspace{-0.2em}
    \caption{\textbf{Comparison of visual token pruning methods at a compression ratio of 50\%.} The value in grey denotes the mean ratio of remaining image features used during generation.}    
        \vspace{-1em}
    \label{tab:ablation}
\end{table}

\noindent \textbf{{Ablation Study.}} We {evaluate} the impact of using attention weights as a {visual feature pruning} criterion. As shown in Table \ref{tab:ablation}, random sampling—where image features are selected randomly—fails to achieve consistent performance across all benchmarks. Additionally, we investigated spatial sampling, a structured approach that selects image tokens in a fixed pattern—every alternate feature index—to approximate holistic image representations. While spatial sampling showed competitive performance on multiple-choice question benchmarks, which we consider less challenging due to the availability of explicit answer choices, it underperformed in more demanding task-driven evaluations (e.g., MME) and open-ended generative tasks (e.g., LLaVA-Bench).

\vspace{-0.5em}

\vspace{+0.05em}
\section{Related Work}
\label{sec:related_work}
\vspace{-0.15em}
\noindent \textbf{LVLMs.} {LLaVA \citep{liu2024visual} and its relevant models \citep{deitke2024molmo, internvl2-8b} combine an LLM with a vision encoder to integrate visual modality features.} Similarly, also using an LLM backbone, the recent LLaMA-3.2-Vision \citep{llama3.2-11b-vision-instruct} {leverages} visual features through cross-attention layers, a design inspired by Flamingo \citep{alayrac2022flamingo}. {This design replaces compute-heavy self-attention layers with cross-modality interaction. We aim to optimize cross-attention-based LVLMs, a relatively under-explored.}

\noindent \textbf{Visual Token Reduction for Efficient LVLMs.} Processing visual features efficiently in LVLMs remains a key challenge. Strategies such as token compression \citep{shang2024llava, cai2024matryoshka} and sparse attention \citep{xiao2023efficient, li2024snapkv, zhang2023h2o, cai2024pyramidkv} optimize visual inputs for the LLM backbones. Examples include FastV \citep{chen2025image}, which exploits sparsity in higher-layer visual attention. ElasticCache and LOOK-M \citep{liu2024efficient, wan2024look}, which merge KV caches to reduce overhead and ZipVL \citep{he2024zipvl}, employing mixed-precision KV caching and importance-based sparse attention for computational gains. However, these advances predominantly target self-attention-based architectures, leaving cross-attention mechanisms underexplored. Moreover, the non-causal relationship between visual inputs and language queries renders direct application of these methods in cross-attention infeasible. Our approach targets effective reducing of cross-attention KV cache without compromising model performance.

\section{Conclusion}
\vspace{-0.15em}

We introduce Trimmed-Llama, a plug-and-play inference optimization method for cross-attention-based LVLMs, leveraging insights from cross-attention weight patterns. By identifying and exploiting inter-layer repetitive cross-attention patterns, our method trims redundant KV caches and reduces computational overhead without additional training. 
\section*{Acknowledgement}
This research was supported by Artificial intelligence industrial convergence cluster development project funded by the Ministry of Science and ICT (MSIT, Korea) \& Gwangju Metropolitan City.

{
    \small
    \bibliographystyle{ieeenat_fullname}
    \bibliography{main}
}

\clearpage
\newpage
\clearpage
\appendix
\section{Appendix}
\subsection{Cross-attention Weight Patterns}
\begin{figure}[hb]
    \vspace{-1.0em}
    \centering
    \includegraphics[width=1.0\columnwidth]{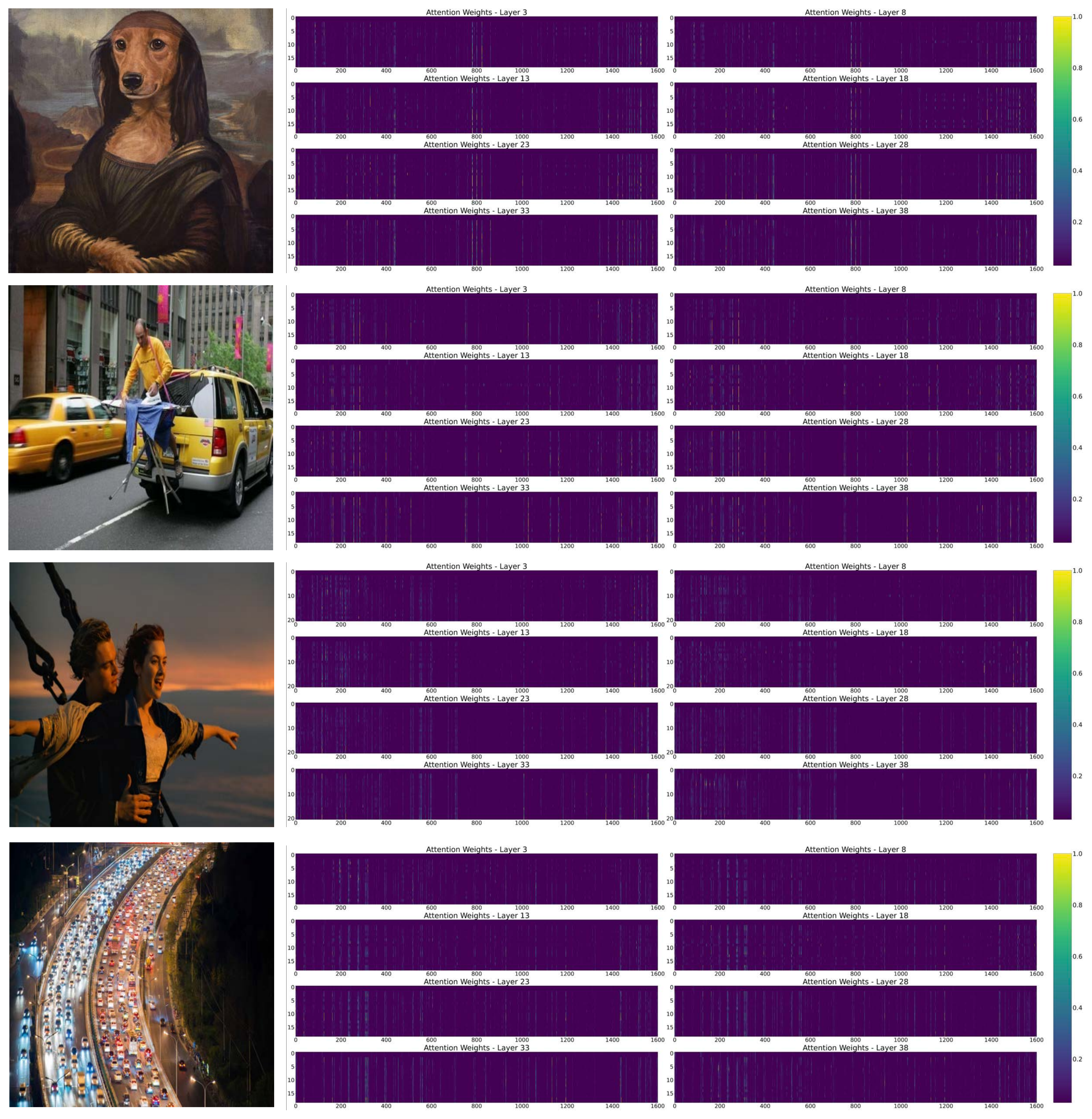}
    \caption{\textbf{Additional results of cross-attention weights.} (Left) Images utilized for the extraction of attention weights. (Right) Cross-attention weight patterns of different layers from corresponding image (x-axis: the index of image features; y-axis: the index of text query features).}
    \label{fig:appx-attn}
\end{figure}

Figure \ref{fig:appx-attn} presents the vertical patterns observed in the cross-attention layers and inter-layer similarities. The attention weights are extracted from samples of LLaVA-Bench's image resized to 384×384 and corresponding instruction, offering a visualization of attention distributions across layers.

\vspace{-1.0em}
\label{appx:attn-pattern}

\begin{figure}[hb]
    \centering
    \includegraphics[width=0.85\columnwidth]{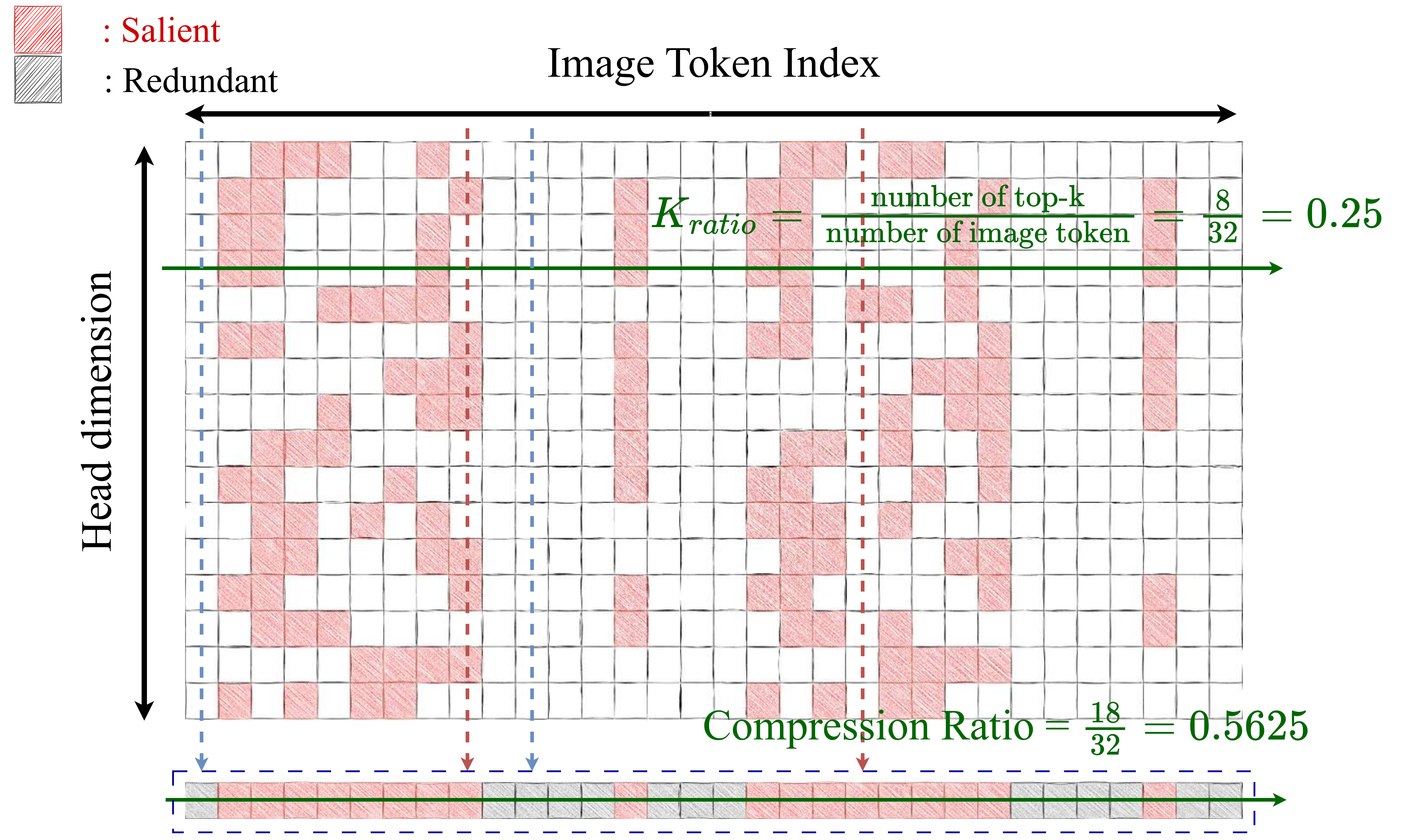}
    \caption{\textbf{Visualization of our token pruning algorithm.} The 2D grid represents image token indices (x-axis) and attention heads (y-axis). The final compression ratio is determined by the fraction of tokens not selected as salient.}
    \label{fig:filter_algo}
\end{figure}
\vspace{-1.0em}

\subsection{Visual Token Pruning Algorithm} \label{appx:filter-algorithm}

Figure \ref{fig:filter_algo} illustrates our proposed algorithm using a conceptual example. Here, the attention map is reduced to two dimensions by summing along the language query dimension. The hyperparameter \(K_{ratio}\) denotes the proportion of salient image tokens retained per attention head. In this example, \(K_{ratio}=0.25\) is used, meaning that each head selects the top-\(k\) (\(k = 0.25 * 32 = 8\)) most attended image tokens.

Based on these selections, an image token is deemed salient if selected by any attention head (indicated by vertical red arrows in the figure), whereas redundant tokens are those not selected by any head (represented by vertical blue arrows). This algorithm effectively captures head-specific token importance, ensuring adaptive attention token filtering across different attention heads.

\begin{figure}[h]
    \centering
    \includegraphics[width=0.65\columnwidth]{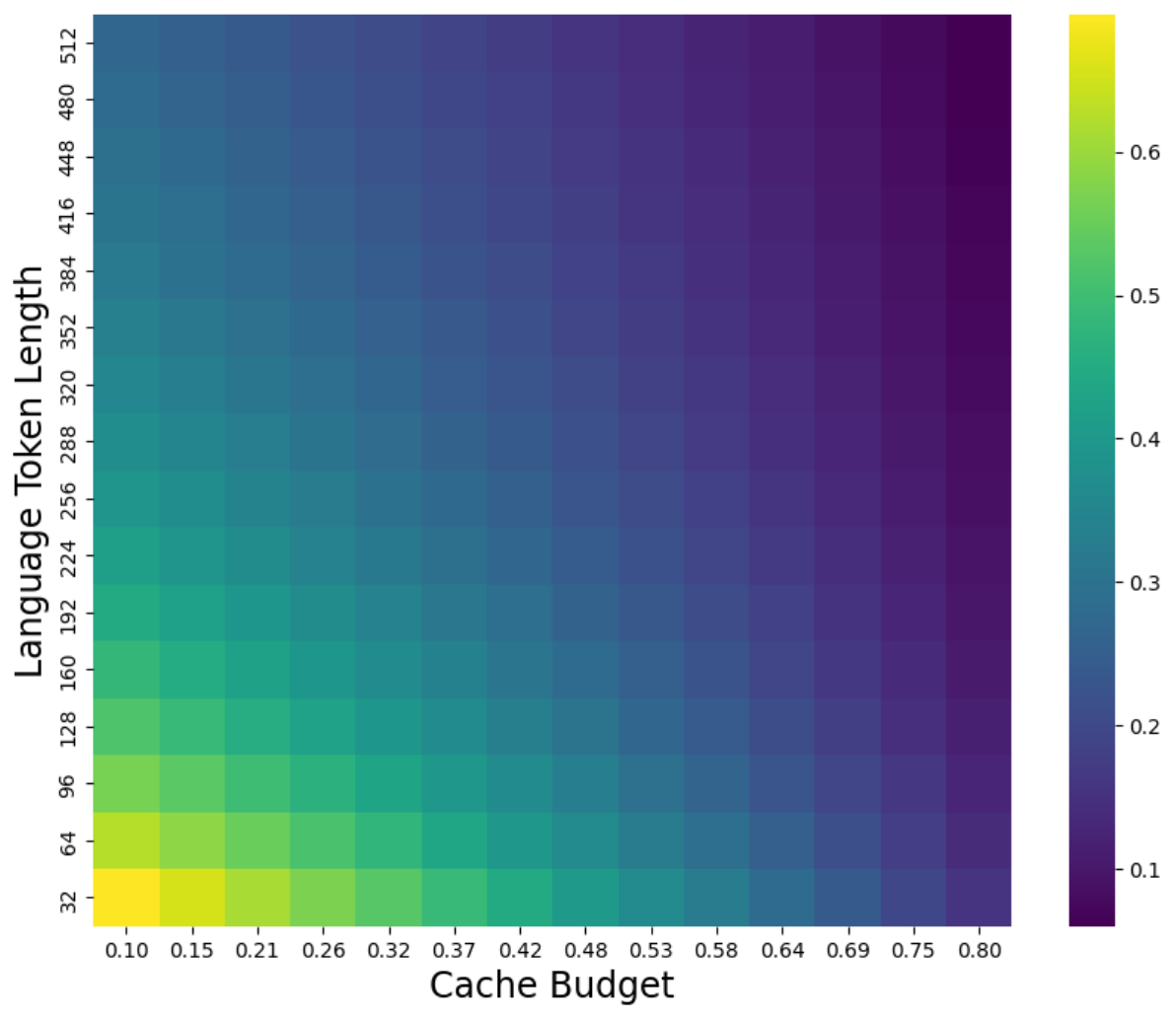}
    \caption{\textbf{Visualization of computational cost reduction.} The heatmap illustrating the theoretical FLOPs reduction ratio is presented with varying KV cache budgets and input sequence lengths.}
    \label{fig:compression_heatmap}
\end{figure}
\vspace{-1.0em}

\subsection{Computational Cost Estimation}
\label{appx:cost-est}


The reduction of computational cost achieved by our approach is presented below. The computation covers multi-head self-attention and cross-attention modules along with feed-forward networks. The image feature is pruned after the first layer with dynamic budget ratio \(R\) produced by the compression method. For estimation, \(n\) denotes the language token length, \(m\) and \(d\) denote the feature dimension of the MLP and attention module, and \(n_k\) denotes the length of the image feature. The number of cross-attention and self-attention layers is indicated as \(C\) and \(S\), respectively.
\vspace{-1.0em}

\begin{gather*} 
    \text{FLOPs}_{self} \colon 4nd^{2} + 2n^{2}d + 2ndm \\
    \text{FLOPs}_{cross} \colon 2nd^2 + 2n_{k}d^2 + 2nn_kd + 2ndm \\
    \text{FLOPs}_{prune} \colon 2nd^2 + 2n_kRd^2 + 2nn_kRd + 2ndm 
\end{gather*}

\vspace{-1.0em}
The theoretical reduction ratio is then calculated as follows. Figure \ref{fig:compression_heatmap} shows a heatmap visualization with different budget ratios \(R\) and input sequence lengths \(n\).

\begin{gather*} 
    1 - \frac{S * \text{FLOPs}_{self} + \text{FLOPs}_{cross} + (C-1) * \text{FLOPs}_{prune}}{S * \text{FLOPs}_{self} + C * \text{FLOPs}_{cross}}
\end{gather*}

\end{document}